\pdfoutput=1

\documentclass[11pt]{article}

\usepackage[preprint]{acl}

\usepackage{times}
\usepackage{latexsym}

\usepackage[T1]{fontenc}

\usepackage[utf8]{inputenc}

\usepackage{microtype}

\usepackage{inconsolata}

\usepackage{times}
\usepackage{latexsym,amssymb}
\usepackage{booktabs}
\usepackage{amsmath}
\usepackage{comment}
\usepackage{graphicx}
\usepackage{stmaryrd}
\usepackage{bm}
\usepackage{dsfont}
\usepackage{xspace}
\usepackage{longtable}
\usepackage{algorithm2e}
\usepackage{multirow,multicol}
\usepackage{lscape}
\usepackage{tabularx}
\usepackage[capitalize]{cleveref}
\usepackage{booktabs}
\usepackage{arydshln}
	
\usepackage{soul}
\usepackage{pdflscape}
\usepackage{afterpage}
\usepackage[utf8]{inputenc}

\usepackage{graphicx}
\usepackage{color-edits}

\usepackage{soul}
\usepackage{listings}
\AtBeginEnvironment{quote}{\texttt\small}

\addauthor{el}{purple}

\title{Language Modeling with Editable External Knowledge}

\author{
 \textbf{Belinda Z. Li\textsuperscript{1}},
 \textbf{Emmy Liu\textsuperscript{2}},
 \textbf{Alexis Ross\textsuperscript{1}},
 \textbf{Abbas Zeitoun\textsuperscript{1}},
\\
 \textbf{Graham Neubig\textsuperscript{2}},
 \textbf{Jacob Andreas\textsuperscript{1}}
\\
\texttt{\{bzl, alexisro, zeitoun, jda\}@mit.edu} \\
\texttt{\{mengyan3, gneubig\}@cs.cmu.edu}
\\
\textsuperscript{1} Massachusetts Institute of Technology, CSAIL \\
\textsuperscript{2} Carnegie Mellon University, Language Technologies Institute
}

\usepackage[disable]{todonotes}
\newcommand{\ourmethod}{\textsc{erase}\xspace}
\newcommand{\ourmethodlong}{\textbf{E}nhancing \textbf{R}etrieval \textbf{A}ugmentation with \textbf{S}elf-consistent \textbf{E}diting\xspace}
\newcommand{\ourdataset}{\textsc{clark}\xspace}
\newcommand{\ourdatasetfull}{\textbf{C}ontinual \textbf{L}earning And \textbf{R}evising \textbf{K}nowledge\xspace}
\newcommand{\bzl}[2][]{\todo[color=red!25, #1]{\textbf{bzl}: #2}}

\newcommand{\gncom}[2][]{\todo[color=green!25, #1]{\textbf{gn}: #2}}

\newcommand{\doc}{d}
\newcommand{\kb}{\mathcal{K}}
\newcommand{\fact}{f}
\newcommand{\facthist}{H}
\newcommand{\plm}{p_\textsubscript{LM}}
\newcommand{\timestamp}{\tau}
\newcommand{\truthval}{v}
\newcommand{\embed}{\mathcal{E}}
\DeclareMathOperator*{\argtopk}{\arg\,\mathrm{top-k}}

\newcommand{\eg}{\emph{e.g.,}\xspace}
\newcommand{\ie}{\emph{i.e.,}\xspace}
\newcommand{\quotetext}[1]{\emph{#1}}
\newcommand{\sect}[1]{\S\ref{#1}}

\begin{document}
\maketitle
\begin{abstract}
When the world changes, so does the text that humans write about it. How do we build language models that can be easily updated to reflect these changes? One popular approach is retrieval-augmented generation, in which new documents are inserted into a knowledge base and retrieved during prediction for downstream tasks.
Most prior work on these systems have focused on improving behavior during \textit{prediction} through better retrieval or reasoning.
This paper introduces \ourmethod, which instead improves model behavior \emph{when new documents are acquired}, by incrementally deleting or rewriting other entries in the knowledge base each time a document is added.
In two new benchmark datasets evaluating models' ability to answer questions about a stream of news articles or conversations, \ourmethod
improves accuracy relative to conventional retrieval-augmented generation by 7--13\% (Mixtral-8x7B) and 6--10\% (Llama-3-8B) absolute.%
\footnote{Code and data are available at \url{https://github.com/belindal/ERASE}}%
\end{abstract}

\begin{figure}[t!]
    \centering \includegraphics[width=0.9\columnwidth]{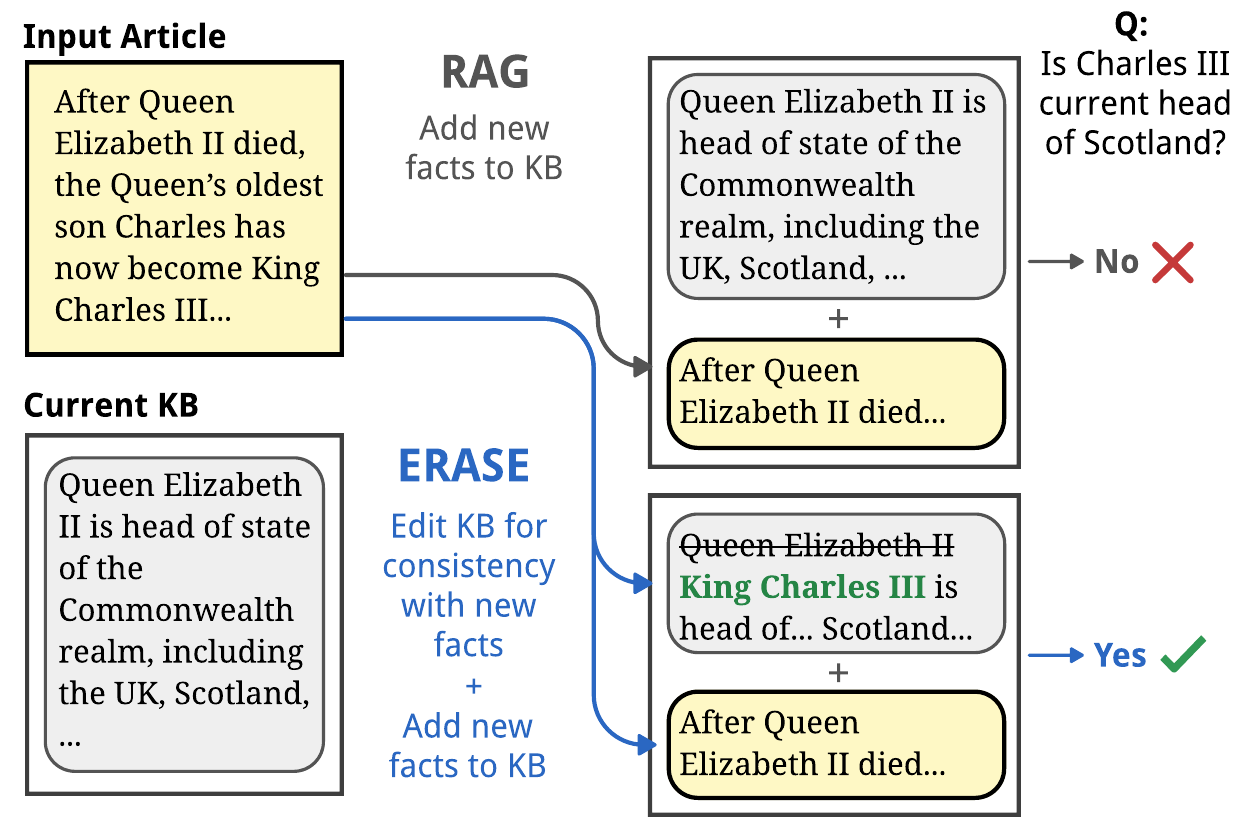}
    \caption{
    In standard retrieval augmented generation (RAG), new facts are simply added to an existing knowledge base $\kb$. This can lead to stale facts in $\kb$, which can in turn lead to incorrect predictions at inference time. In contrast, when \ourmethod reads a new input article, it not only adds new facts to $\kb$, but also \emph{updates} it. \ourmethod can edit or delete (not pictured) existing facts to keep $\kb$ up to date, thereby enabling correct predictions at inference time. The same LM is used to update the memory and make predictions.
    }
    \label{fig:teaser}
\end{figure}
\section{Introduction}
\label{sec:intro}
The world---and the language we used to describe it---are constantly changing.
Consider the example shown in \autoref{fig:teaser}. After reading the article \quotetext{After Queen Elizabeth II died, the Queen's oldest son Charles has now become King Charles III,} a knowledgeable reader might update an entire system of related beliefs, \eg that King Charles III is now also the new head of Scotland. 
How can we train language models and other software systems to reflect these changes?

Continual learning methods tackle the problem of a changing world by incrementally \emph{training} on new information \cite{nell, wang2024comprehensive}. 
But in language models, a simple (and often extremely effective) approach simply presents new information in models' inputs by leveraging either long-context methods
\citep{efficientTansformers} or retrieval augmented generation (RAG; \citealp{RAG}).
which appends new documents to a knowledge base and retrieves a subset of relevant documents to condition on at prediction time \citep{guu2020, lewis2020}.

An important limitation of current RAG approaches is that they sometimes retrieve \emph{stale} documents that have been invalidated by new information. In \cref{fig:teaser}, the article \quotetext{After Queen Elizabeth II died...} would be appended to the existing knowledge base, which includes a fact about Queen Elizabeth's reign when she was alive, \eg \quotetext{Queen Elizabeth II is head of state of...Scotland.} When answering questions about the Scottish head of state, this document might be retrieved, leading the LLM to produce incorrect answers. Past attempts to address this issue have focused on improved \emph{retrieval} methods, but not on ensuring accuracy and consistency of the document collection itself.

This paper describes a method for retrieval-augmented generation that attempts to ensure that the external knowledge base always represents the {current} state of the world. This method, which we call \ourmethod (\ourmethodlong; \sect{s:method}), enables accurate language modeling by updating the knowledge base at \emph{document insertion} time---\ie when new documents are read and added to the knowledge base---rather than at prediction time. Every time a new document is acquired, \ourmethod identifies related documents in the knowledge base and decides whether to keep, edit, or delete them. These operations allow new information to be propagated and prevent stale information from being used for inference. In Figure~\ref{fig:teaser}, \ourmethod not only adds the new article to the knowledge base, but also \emph{edits} the existing fact \quotetext{\st{Queen Elizabeth II} 
$\rightarrow$ \textbf{King Charles III} is head of...Scotland}, thereby enabling correct prediction when this document is retrieved.

We evaluate \ourmethod's performance on question-answering (QA) tasks about a set of continually changing facts described by a stream of text. 
To do so, we introduce a new benchmark dataset, \ourdataset (\ourdatasetfull; \sect{s:dataset}), which contains two domains: (1) \ourdataset-\textsc{News}, a factual QA domain consisting of a set of timestamped news articles paired with questions and timestamped answers; (2) \ourdataset-\textsc{Conversations}, a long-conversation domain where facts about conversation participants evolve over the course of the conversation. The conversation domain contains both single-hop 
and multi-hop edits, the latter of which requires multi-hop inferences at the memory updating stage. 

On this benchmark, \ourmethod outperforms standard RAG baselines and long-context models, giving 
7--13\%
(Mixtral-8x7B) and 
6--10\%
(Llama-3-8B) absolute improvements in accuracy %
compared to standard RAG on the factual QA domain and single-hop section of the conversation domain. On the multi-hop subset, we find that \ourmethod performs comparably to baselines, suggesting there is room for future work to improve multi-hop memory editing.

\section{Background and Related Work}

\ourmethod belongs to a growing body of work aimed at developing LM-based systems that can be updated after training. \ourmethod builds specifically on approaches that update LMs by modifying \emph{inputs} rather than parameters---as discussed below, such methods are more flexible, and often more robust, than alternatives. 

\paragraph{Long-context and retrieval-augmented generation: updating LMs via conditioning} 
One simple and effective way to update LMs is simply to include new information in their context window before inputs to the task of interest (e.g.\ by prepending a question about current events with a sequence of news articles). 
But this approach begins to face challenges when text containing new information is extremely long (e.g.\ comprising thousands of news articles). In these cases, it is neccessary either to use LMs specialized for very long input sequences, or to select a subset of inputs to condition on for each new query to the model (sometimes referred to as retrieval-augmented generation, or RAG). 

Long-context models \citep{wang2020linformer,kitaev2020reformer,press2021train,su2024roformer} focus on modifying LM architectures to allow long sequences to be processed efficiently, or to extrapolate to long inputs.
RAG methods, by contrast, dynamically 
construct relevant contexts tailored to individual queries \citep{guu2020,lewis2020}.
Previous work has explored auxiliary models that selectively choose when to perform retrieval \citep{mitchell2022memory}, or abstain from answering questions when retrieved sources present conflicting or outdated information \citep{chen-etal-2022-rich,zhang-choi-2023-mitigating}.
Other work has examined augmenting LMs with \textit{knowledge graphs} \citep{ijcai2023p0734,modarressi2024memllm}, structured relational knowledge bases that may be timestamped and whose nodes and edges may be updated.
However, such structure can be difficult to construct and risks throwing away essential information; these methods are generally less used than unstructured knowledge bases.

\paragraph{Continual learning: updating LMs via fine-tuning} A broader class of methods, applicable to a much broader class of machine learning models, study the problem of robustly performing \textbf{continual learning} under a non-stationary data distribution \cite{nell, wang2024comprehensive} via training objectives that ensure that new information is retained but old information is not forgotten \cite{jang2022continual,mehta2023dsi, jang2023temporalwiki}. Previous work on LMs has explored the use of continual pretraining \citep{jin-etal-2022-lifelong}, modified pretraining objectives \citep{xu-etal-2023-kilm}, and synthetic data generation \citep{padmanabhan_2023_distill,akyurek2024deductive}.
Continual learning methods are computationally intensive and less widely used than RAG and related methods in language models.

\paragraph{Model editing: updating LMs with targeted interventions} A final category of methods alter LM behavior by making targeted interventions to their parameters, either using specialized secondary ``editing'' models \citep{decao2021editing, mitchell2022fast}
or performing closed-form updates
\citep{meng2022locating, meng2023massediting}. Current methods reliably 
update facts but not all their implications \cite{onoe-etal-2023-lms,hua2024propagation}, and are generally outperformed by retrieval- or fine-tuning-based methods.

\paragraph{Evaluating updates}
Few resources are currently available for evaluating models' ability to generate text about \emph{changing} features of the world while attributing these changes to known source of information.
The Entity Cloze by Date (ECBD) dataset contains entities from Wikidata along with cloze-style sentences \citep{onoe-etal-2022-entity}, and the LoCoMo dataset contains long conversations to measure long-term memory in models \citep{maharana2024lococmo}; unlike \ourdataset, these datasets do not isolate entities whose properties \emph{change} over time.
Many datasets~\citep{zhang-choi-2021-situatedqa,timeqa,meem2024patquestions,dhingra-etal-2022-time,kasai2023realtime,vu2023freshllms} have been released studying temporally-situated question answering; 
however, contexts in these datasets consist only of dates and not source documents.
This makes it difficult to compare results across implementations: were improvements due to a better system, or simply due to a more complete set of documents in the knowledge base?
In \ourdataset, we release both our questions and attributable source documents for those questions.

\begin{figure*}[ht!]
    \centering
    \includegraphics[width=0.9\textwidth]{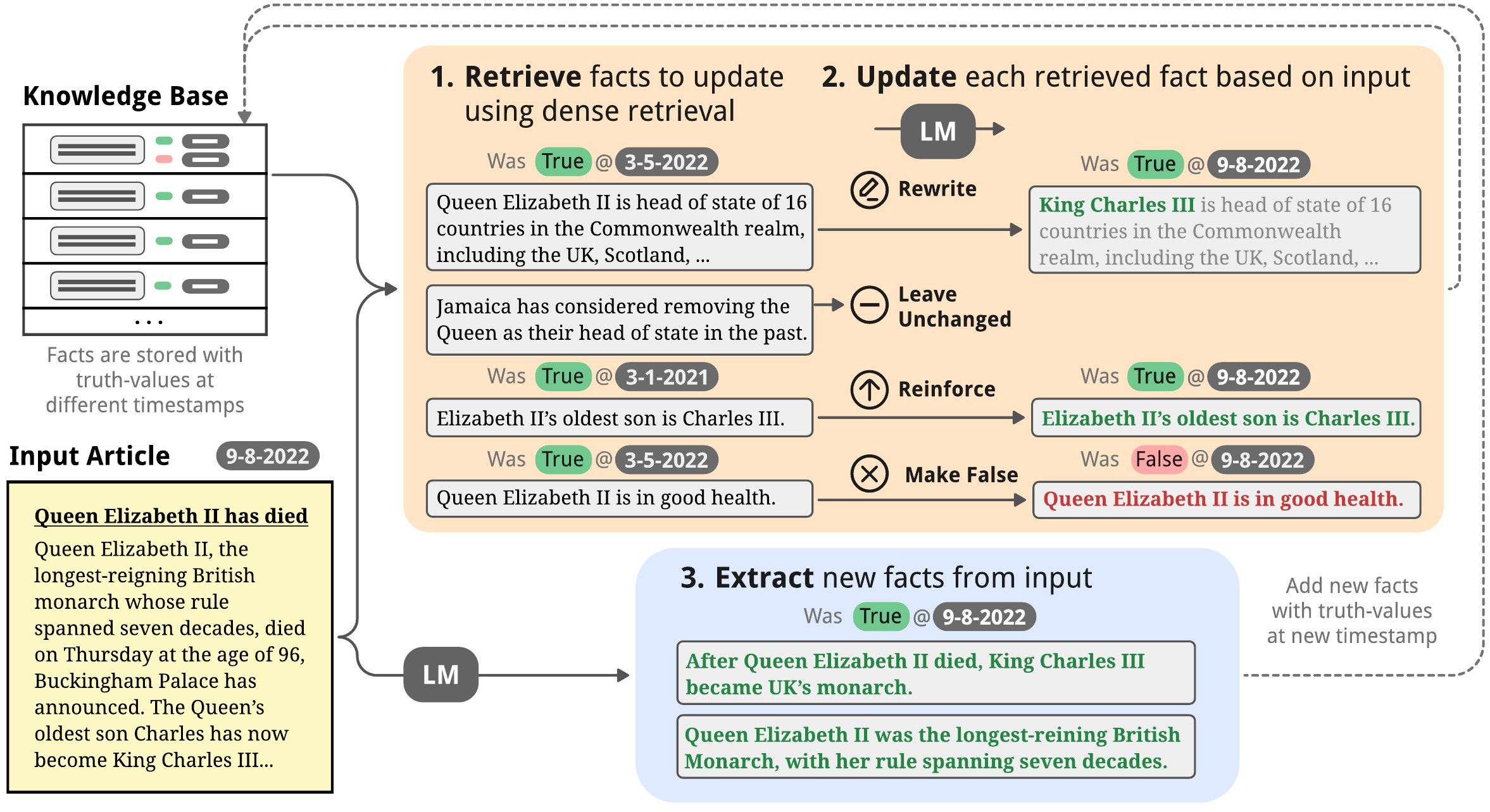}
    \caption{Overview of \ourmethod. We begin by retrieving existing facts relevant to input and prompting a LM to update them. We also extract facts from the input to add to our knowledge base.}
    \label{fig:method}
\end{figure*}

\section{\ourmethod Method}
\label{s:method}

We seek to develop a system that can generate text (e.g.\ for the question answering task depicted in \cref{fig:teaser}) while updating its behavior in response to a continuous stream of documents describing a changing state of the world (e.g.\ the article about the death of Queen Elizabeth II, shown with a yellow background in \cref{fig:method}).
Informally, \ourmethod uses these documents to populate and edit a knowledge base that stores a collection of facts extracted from documents and represented as natural language strings (e.g.\ the identity of the new king, and the duration of Elizabeth II's reign, shown with gray backgrounds in \cref{fig:method}). Importantly, the knowledge base records not just the content of each fact, but when it was first added, and (if relevant) when it ceased to be true. As new documents arrive, \ourmethod attempts to maintain the knowledge base in a \emph{consistent} state---containing only facts that are currently true---by rewriting facts or marking them as false when contradictory facts are introduced by new documents (e.g.\ deleting facts about Elizabeth II's health and updating other references to the UK monarchy). During prediction, \ourmethod then operates like a normal RAG appach: retrieving true facts that are relevant to a given query.

More formally, we begin with a \textbf{language model} encoding a conditional distribution over strings $\plm(\textrm{prediction} \mid \textrm{context})$. 
When a 
new \textbf{document} $\doc_i$ is received with some \textbf{timestamp} $\tau_i$, 
we update the \textbf{knowledge base} $\kb$---each entry in $\kb$ consists of both a \textbf{fact} $\fact_j$ and a \textbf{fact history} $\facthist_j = [(\timestamp_{j0}, \truthval_{j0}), (\timestamp_{j1}, \truthval_{j1}), \ldots]$, where each $\timestamp_{jk}$ is a timestamp and $\truthval_{jk}$ is a \textbf{truth value} indicating whether $\fact_j$ was known to be true or false at time $\timestamp_{jk}$.
We then 
parse the new document into a sequence of facts $\fact_j$ using the LM.\gncom{Separately, is this parsing process described anywhere? It would be nice to explain a little more how we get the facts.} \bzl{yes in the appendix}

Unlike standard RAG methods, it is not in general necessary for facts extracted from documents to correspond one-to-one with facts in the knowledge base: knowledge base entries may also arise by editing old facts in response to new articles. To accomplish this, \ourmethod incorporates new documents into the knowledge base in three steps: \textbf{retrieval}, \textbf{updating}, and \textbf{adding}.

\paragraph{Step 1: Retrieve facts to edit.} 
\begin{equation}
    R \leftarrow \texttt{Retrieve}(\kb, \doc)
\end{equation}
We retrieve a set of knowledge base entries $R = \{(f_{i_0}, H_{i_0}),\cdots (f_{i_m}, H_{i_m})\}\subset K$.
Here we assume that the facts most likely to require \emph{editing} in response to $\doc$ are those most similar to $\doc$.\footnote{For efficiency, we retrieve facts relevant to the entire document in this step, rather than first parsing the document into facts, then retrieving facts relevant to each extracted fact.}
\gncom{This part was a little bit confusing to me, so just to confirm: previous facts are retrieved according to the embedding of $\doc$, not according to the embedding of the facts parsed from $\doc$ (e.g. $f_j$)? With the description up to this point, I would have expected us to try to match new facts $f_j$ with old facts $f_i$ instead of trying to match $\doc$ with $f_i$. It might be nice to add a footnote explaining this design decision for clarity.}
Following most modern RAG approaches~\citep{RAG}, \ourmethod performs \textbf{dense vector retrieval}, using a learned embedding model $\embed$ to assign documents and facts vector representations, then retrieve a set of $m$ to optimize:
\begin{equation}
    \texttt{Retrieve}(\kb, \doc) = \argtopk_{(\fact_j, \facthist_j) \in \kb} ~\embed(\doc)^\top \embed(\fact_j) ~ .
\end{equation}

\paragraph{Step 2: Update retrieved facts.} 
\begin{align}
    &\forall (\fact_j, \facthist_j) \in R, \, (\fact'_j, \facthist'_j) \leftarrow \texttt{Update}(\fact_j,\facthist_j,\doc,\timestamp) \nonumber \\
    & \kb \gets \kb \cup \{(\fact'_j, \facthist'_j)\}
\end{align}
We update the knowledge base by modifying each retrieved fact $f_i\in R$ in one of the following ways:
\begin{itemize}
    \item \textbf{Reinforce fact}: If the fact $\fact$ is supported by $\doc$, we add $(\texttt{true}, \timestamp)$ to $\facthist$. An example of such a case would be $\fact =$ \textit{Mary works in a warehouse} and $\doc =$ \textit{Mary came back from her job at UPS where she loaded and sorted packages all day}.
    
    \item \textbf{Keep fact unchanged}: If $\doc$ is irrelevant to $\fact$ or does not affect the truth value of $\fact$, then we do nothing and let $\fact' = \fact$ and $\facthist' = \facthist$. An example of such a case would be $\fact =$ \textit{Mary works in a warehouse} and $\fact = $ \textit{Mary took a jog in the park}.

    \item \textbf{Make fact false}: If $\fact$ is contradicted by $\doc$, we add $(\texttt{false}, \tau)$ to $\facthist'$. An example of such a case would be $\fact =$ \textit{Mary works in a warehouse} and $\doc = $ \textit{Mary got fired from her warehouse job}.

    \item \textbf{Rewriting}: Alternatively, if $\fact$ is contradicted by $\doc$, we may \textit{rewrite} it into a new expression $\fact'$ that is inferrably true from $\doc$ and the subset of retrieved facts $\subset R$ that have been \textit{reinforced} or \textit{kept unchanged}. %
    We then replace the old KB entry $(\fact, H)$ with a new KB entry $(\fact', [(\texttt{true},\tau)])$.
\end{itemize}

For all operations above, we prompt an LM (which may be the same LM used for prediction) to classify each retrieved fact into one of \textit{reinforce, no change, make false}.\footnote{The task in the first pass is similar to a fuzzy version of natural language inference classification. Inputs that make facts more likely (even if they do not exactly entail those facts) are classified as \textit{support}, and inputs that make facts less likely (even if they do not exactly contradict those facts) are classified as \textit{make false}.}
We then iterate through all facts classified as \textit{make false}, and ask the LM if it can rewrite the fact into a true expression. In this second phase, the LM is allowed to condition on facts that it classified as \textit{reinforce} or \textit{no change}, allowing it to potentially handle multi-hop edits. The full details of this procedure can be found in~\cref{app:update_prompts}.

\paragraph{Step 3: Add new facts.} 

\begin{equation}
    \kb \leftarrow \kb \cup \texttt{Add\_facts}(T)
\end{equation}
We add all new facts by conditioning on $\doc$ and prompting the LM to extract atomic facts $\fact$. The prompt we use can be found in~\Cref{app:fact_extraction}.
Analogously, ~\citet{chen2023dense} used a \textit{propositionizer} to decompose articles into propositions.

\paragraph{Prediction:}
To use an \ourmethod system after updating, generation is performed using a standard RAG pipeline described in step 1.
We condition on both the retrieved facts and their corresponding history in context. The full prompt can be found in~\Cref{app:infer_prompt}.

\begin{figure*}
    \centering
    \includegraphics[width=\linewidth,trim={0 0 0 0},clip]{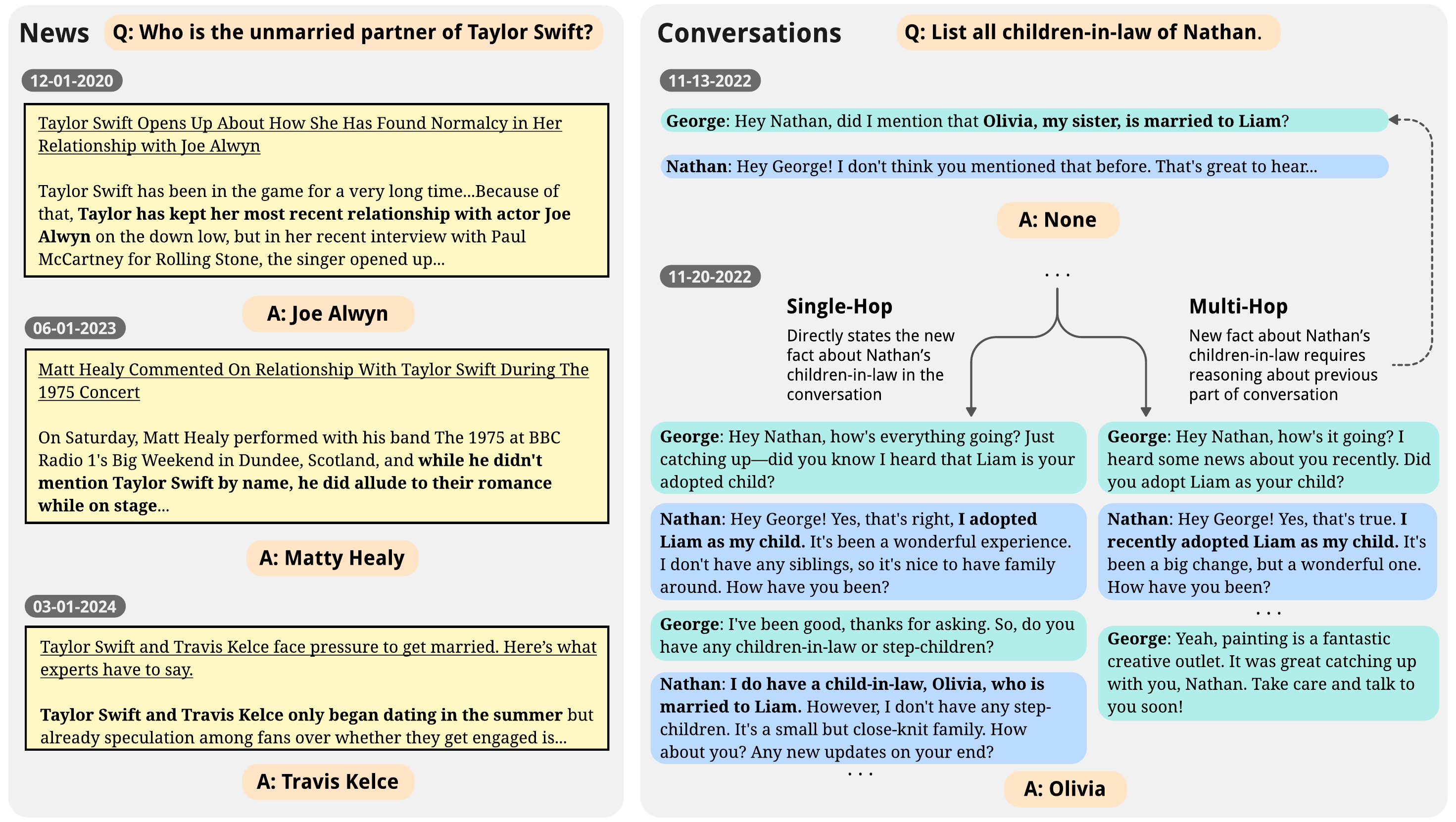}
    \caption{Sample data from our datasets. The News dataset consists of factual questions whose answers change over time, with the associated source inducing that change. The Conversations dataset consists of conversations between two personas with evolving life facts. The single-hop subset directly states all facts that are changed, while the multi-hop subset requires reasoning about previous chunks of conversation to infer all changes.}
    \label{fig:dataset}
\end{figure*}

\section{Dataset}
\label{s:dataset}

We construct two datasets to evaluate \ourmethod.
We acquire a set of natural-language texts $L_t$, a set of ground truth world states $W_t$ and a series of questions $q_0\cdots q_n$ associated with $W_t$.
We focus on questions that \textit{update} over time: the set of questions we ask at each timestep are the same, but each question is associated with a list of timestamped answers $(q_i, \{(a_{i0}, t_{i0}), (a_{i1}, t_{i1}),\cdots\})$. %
The datasets span two domains where continual learning is useful: one about the evolving state of the world, and one about the evolving state of agents in a conversation. Samples from each dataset can be found in~\Cref{fig:dataset}. An overview of state transitions and questions in these two datasets can be found in \autoref{app:dataset_stats}.

\subsection{News Articles}

\paragraph{World States}
In this domain, world states are expressed in the form \texttt{(subj, rel, obj)}: for instance, \texttt{(Elizabeth II, position held, monarch of the United Kingdom)}. We mine these triples from Wikidata.\footnote{\url{https://www.wikidata.org/}, which is public domain. Its license can be found at \url{https://www.wikidata.org/wiki/Wikidata:Licensing}.} As Wikidata is updated over time, each fact is also associated with a start and end date. To find changed facts, we extract \texttt{(subj, rel)} pairs for which there are at least two distinct fact relations at different timestamps between November 2021 and April 2024. Through this process, we obtain 1,174 triples for 10 unique relations, summarized in \Cref{tab:news_relations}.

\paragraph{Documents}
For each world state \texttt{(subj, rel, obj, start\_ts, end\_ts)}, where the start and end timestamps are extracted from Wikidata, we obtain an English article confirming that fact between the start and end timestamps, validated by crowd workers. Through this process, annotators collected a total of \textbf{1149} articles.\footnote{Note 1149 $<$ 1174, meaning at least a few articles were shared across relations -- these represent difficult cases where a single article makes multiple relation changes.} See \Cref{app:wikidata} for details. These documents---rather than raw relation triples---are the input to \ourmethod.

\paragraph{Questions and Answers}
We automate the generation of questions and answers from $W$ by writing templates for each relation and generating questions and answers from those templates. We generated a total of \textbf{1409} questions. The full list of templates can be found in~\Cref{app:wikidata}.

\subsection{Synthetic Conversations}
Following prior work~\cite{maharana2024lococmo}, we construct a synthetic conversation domain by placing two LLMs with different personas in conversation with each other. Conversations are engineered to reflect changing facts in the agents' simulated lives. A detailed overview of dataset construction can be found in \cref{app:convos}.  To validate the LM generations, three authors manually examined 3 conversations (1008 questions) in total and got an average of 95\% accuracy on these questions.

This synthetic domain allows us to rigorously control and evaluate forms of reasoning that may be hard to isolate in natural data like news articles. 

\paragraph{World States}
We generate an independent world for each conversation.
We model the world underlying a conversation as a Markov chain with states $S$, described by a list of \texttt{(subj, rel, obj)} relations, and allowable transitions $T(S)$. States $S$ are defined by entities including people, companies, jobs, hobbies, along with mutable and immutable relations between them.
Transitions $t\in T(S)$
change one or more relation in the state: for example, \textit{Bob changed jobs to work at Google} changes the \textit{employees} of Google, the set of \textit{coworkers} of Bob, the set of \textit{coworkers} of all Google employees, and the set of \textit{coworkers} of all employees of Bob's former company, etc. At each timestep, we sample a transition from $T(S)$ uniformly at random.
The full list of entities, relations, and transitions and their downstream effects can be found in~\Cref{app:convos}.

\paragraph{Conversations}
We generate conversations by sampling two people in the world $p_1$ and $p_2$ and prompting two LLMs with their corresponding personas and the initial world state $S$.
We then generate twelve conversation ``chunks''---separated by time---by sampling state transitions between \textit{every other} 
chunk and having people converse about the facts that have changed after each transitions.

We also construct a challenge set of \textit{multi-hop} updates in this domain, which require propagating changes to multiple downstream facts and reasoning about global coherence between facts. For example, Bob may mention that he has changed his job but may not mention that \textit{Jane is no longer his coworker} or that \textit{Mary (who works at Google) is now his coworker}. The LM must make multi-hop inferences to update the latter two facts.

We generate \textbf{100} conversations (50 single-hop, 50 multi-hop) in total. Conversations were on average \textbf{11045} tokens long in the single-hop subset and \textbf{11069} tokens long in the multi-hop subset. Detailed statistics may be found in Appendix \Cref{fig:convos_stats}.

\paragraph{Questions and Answers}
Given a world state at time $t$, we query \textit{all} facts about the world. Similar to the news setting, we automate generation of questions and answers through templates. We generate \textbf{140} questions per conversation.

\begin{figure}[t!]
    \centering
    \includegraphics[width=.8\linewidth,trim={1cm 1cm 13.5cm 1cm},clip]{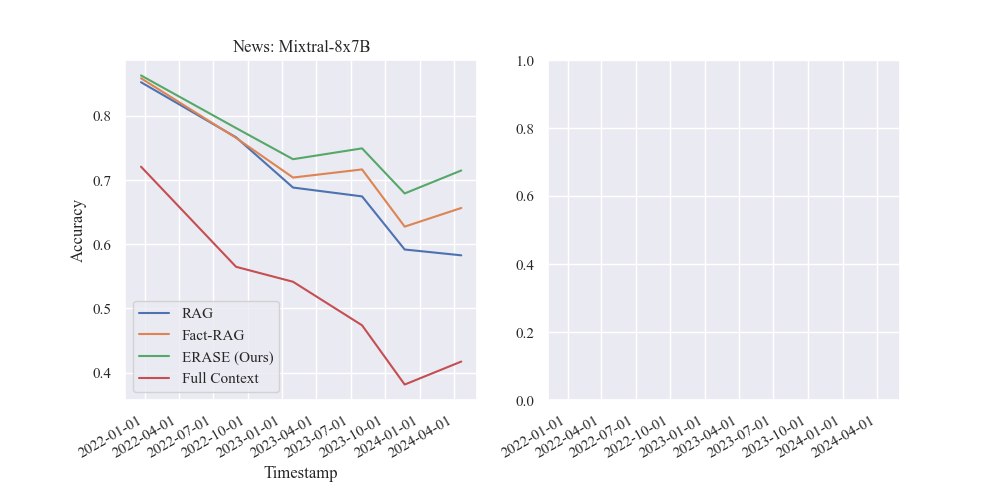} \\
    \includegraphics[width=.8\linewidth,trim={1cm 0.5cm 13.5cm 1cm},clip]{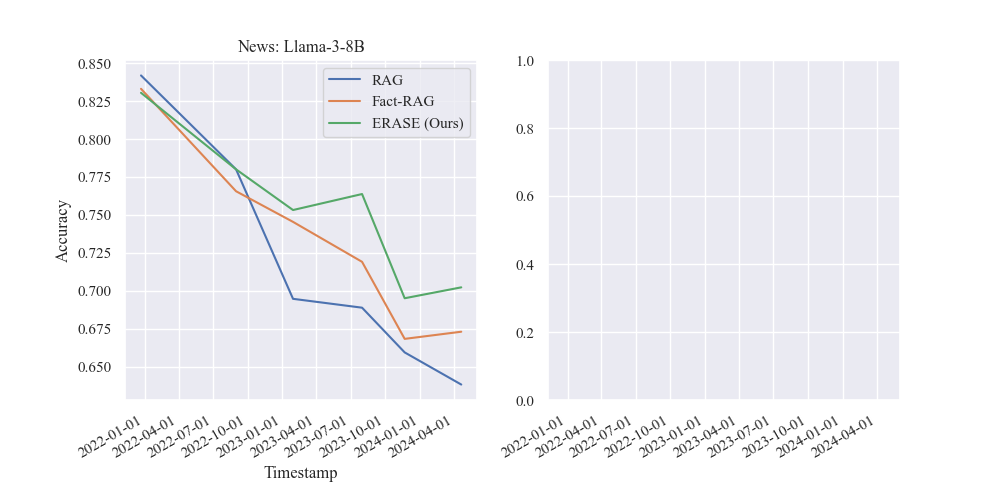}
    \caption{Mixtral-8x7B (top) and Llama-3-8B (bottom) results on the news article domain. %
    \ourmethod outperforms RAG, RAG with fact-level granularity, and even long-context models, especially in later timesteps as more new information is learned.}
    \label{fig:wiki_results}
\end{figure}

\section{Experiments}
\label{s:experiments}

In our experiments, we present to a LM articles or conversational turns in chronological order, and periodically ask questions about the state of the world (as described by input documents) at that point in time.

\subsection{Evaluation and Metrics}
\paragraph{News articles}
We present the model with a stream of articles
ordered by timestamp. As all answers are dated with a start and end timestamp, we always know which answer is true for a given timestamp.\footnote{Note that this does not correspond to when these facts became true and false in the real world, but rather to when the article introducing the changed fact was written and read.}
We ask questions at regular intervals, at timesteps corresponding to when 20\%, 40\%, 60\%, 80\%, and 100\% of the total world state changes have been revealed to the model.
Because it is too expensive to ask every question at every timestep, we ask \textit{all questions whose answers have changed} $Q$, then sample a subset of \textit{questions whose answers have not changed} $Q'$, such that $|Q'| = |Q|$.
We design each question as a multiple choice question, where the model is asked to select between all answers that have been true for the question in the past, present, or future.
This ensures that the negative options are sufficiently difficult, and allows us to probe for the models' updating capabilities.
We report exact-match accuracies between the model-predicted answer to the true answer.

\paragraph{Conversation} We evaluate each conversation independently, and report the mean and standard error of scores over each conversation.
We stream in \textit{chunks} of conversations into the model, and 
ask questions after each conversation chunk. Similarly to the news domain, we subsample questions whose answers have not changed, such that at each timestep we are asking the same number of questions whose answers have changed as those whose answers haven't changed. 
For questions that have multiple true answers (e.g. \texttt{List all siblings of Liam}), we measure the set equality between the generated and true sets of answers.
Otherwise, we use the same exact match accuracy as we use for the news articles domain.

\begin{table*}[t!]
    \centering
    \footnotesize
    \begin{tabular}{|p{1cm}|l|ccc|ccc|}
    \hline
        & & \multicolumn{6}{c|}{Data Subset} \\
        & & \multicolumn{3}{c}{Single-hop} & \multicolumn{3}{c|}{Multi-hop} \\
         & & 0 updates & 1 update & \multicolumn{1}{c}{2+ updates} & 0 updates & 1 update & 2+ updates \\
    \hline
        \multirow{4}{1cm}{Mixtral-8x7B} 
        & RAG \cite{RAG} & $\mathbf{86.0_{\pm 0.7}}$ & $56.7_{\pm 1.8}$ & $50.9_{\pm 3.2}$ & $\mathbf{84.5_{\pm 0.8}}$ & $\mathbf{20.9_{\pm 1.4}}$ & $20.0_{\pm 2.3}$ \\
        & Fact-RAG \cite{chen2023dense} & $82.7_{\pm 0.8}$ & $51.5_{\pm 1.8}$ & $52.7_{\pm 3.1}$ & $81.8_{\pm 0.8}$ & $18.0_{\pm 1.3}$ & $\mathbf{30.2_{\pm 2.7}}$ \\
        & \ourmethod (Ours) & $82.0_{\pm 0.8}$ & $\mathbf{59.1_{\pm 1.8}}$& $\mathbf{57.9_{\pm 3.1}}$ & $81.5_{\pm 0.8}$ & $\mathbf{20.1_{\pm 1.4}}$ & $27.2_{\pm 2.6}$ \\
        \cdashline{2-8}
        & Full Context & $88.8_{\pm 0.6}$ & $71.6_{\pm 1.6}$ & $75.7_{\pm 2.4}$ & $88.4_{\pm 0.6}$ & $43.2_{\pm 1.7}$ & $54.3_{\pm 2.8}$ \\
    \hline
        \multirow{3}{1cm}{Llama-3-8B} & RAG \cite{RAG} & $\mathbf{84.4 _{\pm 0.7}}$ & $57.8_{\pm 1.8}$ & $55.2_{\pm 3.1}$& $\mathbf{83.6_{\pm 0.8}}$ & $22.2_{\pm 0.1}$ & $26.8_{\pm 2.6}$  \\
        & Fact-RAG \cite{chen2023dense} & $82.6_{\pm 0.8}$ & $62.6_{\pm 1.7}$ & $62.0_{\pm 3.0}$ & $81.2_{\pm 0.8}$ & $\mathbf{26.4_{\pm 1.6}}$ & $\mathbf{32.1_{\pm 2.8}}$ \\
        & \ourmethod (Ours) & $82.0_{\pm 0.8}$ & $\mathbf{65.3_{\pm 1.7}}$ & $\mathbf{65.2_{\pm 2.9}}$ & $81.0_{\pm 0.8}$ & $\mathbf{26.5_{\pm 0.2}}$ & $\mathbf{31.7_{\pm 2.7}}$ \\
    \hline
    \end{tabular}
    \caption{Results on the synthetic conversation domain. Full context serves as a skyline in this domain as the full conversation fits into the context window. We compare against other retrieval-based methods. In \textbf{bold} are results that are the \textbf{statistically significantly best} out of all other methods in the same setting (model, data subset, \# updates). While \ourmethod significantly improves single-hop edits in both models, it still struggles with multi-hop edits. Small LMs make errors in multi-hop reasoning during the overwriting stage, and suspect that as LMs improve multi-hop reasoning, we will see greater gains with \ourmethod. \\
    \footnotesize{* We merge 2+ updates as generally there is a long tail of questions with more updates. Only 27 questions total have 3+ updates.} 
    }
    \label{tab:convo_results}
\end{table*}

\subsection{Models}
We use a Mixtral 8x7b Instruct model (56B parameters; \citealp{jiang2024mixtral}), queried using Together AI\footnote{\url{https://www.together.ai/}}, and a local copy of Meta's Llama-3 8b Instruct model (8B parameters ; \citealp{llama3modelcard}) run on one NVIDIA A100 GPU.\footnote{Llama-3 8b has knowledge cutoff of March 2023. Mixtral's has not been published, but appears to be around late 2022 or early 2023.} 
For all prompts during inference and update-time, we sample from the LM with temperature 0.
We use GTR (T5-large; 770M parameters; \citealp{ni-etal-2022-large}) as $\embed$ to encode queries and documents for dense retrieval, both in the inference stage and the retrieval step of updating. We use %
a fast inner-product search datastructure for efficient retrieval~\citep{FAISS}. 
For prompting during the updating stage, we use the same LM that we are using for inference.
We restrict the context window to 4096 for the news domain and 2048 for the conversation domain.\footnote{Note this is smaller than the original context windows for these models, both to run our experiments efficiently, and to test out a (realistic) scenario where the total number of new world changes cannot fit into the context window of a language model.}
Inference and updating took a few hours to complete for both models and for all method.
At inference time, we allow all models to perform zero-shot chain-of-thought, giving them an additional ability to reason about inconsistent facts at inference time. 

\subsection{Baselines}
We compare \ourmethod to three baselines:

\paragraph{RAG} RAG~\cite{RAG} stores and retrieves text at the granularity of \textit{passages}. We save each article and conversation chunk as a separate passage in the knowledge base. For long articles and conversation chunks, we divide them into passages of length \verb|context_window / 2|.

\paragraph{Fact-RAG} To isolate the effects of \textit{editing}, we benchmark against a version of RAG that stores and retrieves \textit{facts} in the knowledge base, akin to~\citet{chen2023dense}. 
We implement this baseline by prompting LMs to extract facts from passages, i.e.\ step 3 of \ourmethod, which 
outperformed the propositionizer from~\citet{chen2023dense}.

\paragraph{Long context LMs} Mixtral-8x7B has a long context window of 32k. 
We run an in-context learning baseline %
by conditioning Mixtral %
on all news articles or conversation chunks, presented in chronological order. These texts are timestamped, and Mixtral is able to condition on the most recent set of texts up to its context limit when making predictions. In the Conversations domain, this condition serves as a skyline since conversations fit completely into the context window.

\section{Results}
\label{sec:results}
\Cref{fig:wiki_results} and \Cref{tab:convo_results} show results for the news and conversation domains respectively.

\paragraph{\ourmethod improves over standard RAG with passage retrieval.} For both Mixtral and Llama-3 in both domains, we see significant improvements using \ourmethod over RAG, particuarly as the number of edits increases. For example, in the news domain, at the final timestamp after reading all articles, Mixtral with \ourmethod is 13 points better than Mixtral with RAG, while Llama with \ourmethod is about 6 points better than Llama with RAG.
We see similar trends on the single-hop subset of the conversation domain: for questions with 2+ updates, \ourmethod is 7 and 10 points better than RAG, using Mixtral and Llama respectively.

\paragraph{Editing existing facts improves beyond RAG with fact retrieval.}
For both Mixtral and Llama-3, \ourmethod substantially improves performance over Fact-RAG as the number of edits increases, on both the news domain and the single-hop subset of the conversation domain.
Improving knowledge base consistency helps, \textit{even with step-by-step reasoning} at inference-time. %

\paragraph{In the news domain, \ourmethod improves over long-context modeling.} 
In~\Cref{fig:wiki_results}, we plot Mixtral with its full context window on the news domain. Long-context models are unable to scale as more articles are added. However, we find that \ourmethod (and retrieval methods generally) are unable to compete against fitting full conversations in the context window~\Cref{tab:convo_results}. That said, the cost of conditioning on full conversations is greater than the cost of conditioning on simply retrieved facts, especially as the number of queries per conversation increases.\footnote{Conditioning Mixtral on full conversations costs 7.3K tokens per query, whereas retrieval costs $\sim$ 1.7K tokens per query $+$ a fixed cost of $\sim$ 42k tokens per conversation chunk. Generally in the real world that the number of queries far outflanks the number of documents generated about changes in the world. In our dataset without subsampling, full context would cost 102M tokens while ours would cost 28M tokens.}

\paragraph{Multi-hop retrieval and editing is still challenging.}
Both LMs struggle with the multi-hop subset of the conversation dataset.
We believe this isn't a drawback of fact editing itself, but of our implementation of it: a qualitative examination of failure cases (see \Cref{sec:multihop_errors} for some examples) revealed that 
our retrieval model often failed to retrieve all downstream facts that need to be edited,
and language models on the scale of Mixtral-8x7b and Llama-3-8b struggled with reasoning about multi-hop edits, failing to make those edits when necessary.
A more powerful retrieval and editing model may be able to avoid these errors.

\section{Conclusion}
This paper introduced \ourmethod, an approach for \textit{editing existing facts} in a knowledge base when new documents are being inserted. We also introduced two datasets for testing the ability of models to update their knowledge, accompanied by documents that induce those changes. Editing existing facts brings significant improvements to RAG-based models. Even if future models become better at reasoning about inconsistencies with scale, fact editing is useful for amortizing the cost of reasoning about consistency \textit{at insertion time}, rather than having to re-evaluate consistency each time a fact is queried.
Future work can focus on improving any part of the update pipeline, particularly focusing on retrieving downstream facts (step 1) that will be affected by an input (which is different from retrieving simply \textit{relevant} facts), and improving LM ability to perform multi-hop updates (step 2).

\subsubsection*{Limitations}
As noted in ~\Cref{sec:results}, \ourmethod is still subpar for multi-hop updates, largely due to retrieval model's inability to retrieve all the necessary facts and the LMs' inability to reason about multi-hop edits. We believe that this limitation can be mitigated with better retrieval models and better LMs.

Second, because LMs have a tendency to hallucinate, allowing LMs to directly edit the knowledge base may introduce noise into the knowledge base. While our results found that the utility of propagation was greater than any hindrance due to such noise, this noise has the potential to snowball on long timescales as the number of new passages and edits grows beyond tens of thousands, hundreds of thousands, or millions.
That said, we do not believe this limitation is inherent to knowledge-base editing: future work can explore more principled and rigorous approaches to editing with guarantees around what edits are made and to how many facts.
Furthermore, we believe that for any approach to model editing, there is a natural tradeoff between noise and edit coverage. 

Finally, having to process each document and update the knowledge base is less efficient than simply adding it to the retrieval store. We justify this cost by assuming that the number of insertions is far fewer than the number of queries. (For example, Forbes reports that 252,000 websites are created per day,\footnote{\url{https://www.forbes.com/advisor/business/software/website-statistics/}} while Google receives about 8.5 billion searches daily.\footnote{\url{https://seo.ai/blog/how-many-people-use-google}}) Thus, by shifting the cost of reasoning about consistency from query-time to insertion-time, \ourmethod is arguably \textit{more efficient} in practice than RAG.

\subsubsection*{Ethical Considerations}
Being able to interpretably edit models is useful for improving the safety and trustworthiness of models. If there is misinformation in the knowledge base, our method allows these facts to be corrected quickly and these corrections to propagate through the knowledge base.
Our method magnifies the effect of each change, making it easy for system designers to keep knowledge up-to-date and remove any stale or incorrect knowledge.
Conversely however, this could also empower malicious actors to insert false facts, which will also be propagated through the knowledge base.
There will need to be safeguards in place to ensure that any inserted and propagated knowledge is from reliable sources, with potential vetting of each inserted article. One of the pros of \ourmethod is that we can see every LM operation occurring in real time: any update operation can be examined manually to ensure that the changes are desirable.

\bibliography{custom}

\newpage

\appendix

\section{Prompts for \ourmethod}
\label{app:prompts}
In this section, we list all prompts that we use for each step of our method.

\subsection{Fact Updating}
\label{app:update_prompts}
In practice, we implement these operations by performing \textit{two passes} over the retrieved facts.\bzl{TODO potentially abstract away this detail} In the first pass, we prompt the LM with the input $\doc$ and each fact $\fact\in R$ and prompt it to \textit{classify} the fact into one of \textit{reinforce, no change, make false}.
From this first pass, we divide the retrieved facts into two sets: $R_\text{true}$, comprising facts that remain true (\textit{reinforce, no change}), and $R_\text{false}$, comprised of facts that have become false (\textit{make false}).
In the second pass, we iterate through $R_\text{false}$, and prompt the LM to rewrite the fact into a true fact (if possible), conditioned on the new document $\doc$ and $R_\text{true}$. This serves a few purposes:
\begin{enumerate}
\item If $\fact$ is only made partially false by $\doc$, we may retain information expressed in $\fact$ but not $\doc$. For example, if $f$ is \textit{Mary and Bob work at UPS}, and $\doc$ is \textit{Mary got fired from UPS}, we may rewrite $\fact$ as \textit{Bob works at UPS}, rather than negating the entire fact.
\item Conditioning on $R_\text{true}$ allows the LM to make \textit{multi-hop} edits. For example, if $\fact$ is \textit{Mary is coworkers with Bob}, and $\doc$ is \textit{Mary changed workplaces to Amazon}, if $R_\text{true}$ contains \textit{Quinn works at Amazon}, then we can rewrite $\fact$ as \textit{Mary is coworkers with Quinn}.
\end{enumerate}

First round: classifying facts as becoming more or less likely to be true. 
\begin{lstlisting}
[Input] [Timestamp: {ts}] {context} [End Input]

The fact "{fact}" was previously true. In light of the input, is "{fact}" likely still true as of {ts}? Begin by summarizing the changes we learned from the input, then reasoning briefly about them to give your final answer with "Answer: Reinforce" (if the input makes the fact more likely) or "Answer: Make False" (if the input makes the fact less likely) or "Answer: No Change" (if the input doesn't affect the fact, e.g. if the input is irrelevant to the fact). Assume that the fact is still true (keep true) if nothing in the input contradicts it.
\end{lstlisting}

Second round: extracting rewrites
\begin{lstlisting}
[Input] [Timestamp: {ts}] {context}
Other True Facts at {ts}: {", ".join(still_true_facts)}
[End Input]

The fact "{fact}" was previously true but no longer. Given the above input and true facts, can you rewrite it into one that is true as of {ts}? Output your answer in form "rewrite: rewritten fact" or "no rewrite possible".
\end{lstlisting}

\subsection{Fact Extraction}
\label{app:fact_extraction}
\begin{lstlisting}
Extract all facts from the input text, with each fact on a new line and without bullet points or numbered lists. Facts should be simple, independent, standalone, and decontextualized. Break up long facts into smaller facts. Resolve all references (e.g. pronouns, definite articles, etc.) by copying full reference object everywhere it is referenced. Only include facts referring to the current world state (what is true *now*), as opposed to facts true in the past. If there are no facts, please output "No new facts." Do not include any other text.
\end{lstlisting}

\subsection{Inference}
\label{app:infer_prompt}
Given a question \texttt{question} at timestep \texttt{ts} (and choices \texttt{answer\_choices}), 
We first retrieve facts $\fact_i, [(\tau_{i0}, v_{i0}), (\tau_{i1}, v_{i1}), \cdots]$ from the knowledge base with similarity threshold $> 0.7$ to \texttt{question}. We then prompt a LM with the following: 
\begin{lstlisting}
Read the statements/passages below then answer the question below

***BEGIN STATEMENTS***
{f_i} ({v_{i0}} at {tau_{i0}}, {v_{i1}} at {tau_{i1}}, ...)
{f_j} ({v_{j0}} at {tau_{j0}}, {v_{j1}} at {tau_{j1}}, ...)
...
***END STATEMENTS***

Given the above statements are true and any prior knowledge you have, answer the following question at timestep {ts}?:
{question}

Briefly reason then answer with one of: {answer_choices}.
\end{lstlisting}

For questions requiring list answers (e.g. list all the siblings of Rachel), we replace the last line with:
\begin{lstlisting}
Briefly reason then answer with a JSON list, ["item1", "item2", ...], of zero or more of the following items: {answer_choices}. If you include any of the above items, make sure to copy their names exactly as is from the list. Your list may be empty, [], if none of the answers are true.
\end{lstlisting}

\section{Dataset Construction Details}
\subsection{News Articles}
\label{app:wikidata}

We construct this dataset in three stages:
\paragraph{Extracting World States $W$.} We retrieve \texttt{(subj,rel)} pairs from Wikidata for which there are at least two distinct fact relations at different timestamps, e.g. \verb|(subj,rel,obj1,start_ts1,end_ts1)| and \verb|(subj,rel,obj2,start_ts2,end_ts2)|. These timestamped facts are used to ``represent'' $W$. We filter for subjects \texttt{subj} located in English-speaking countries to ensure we can find English-language sources.
We use SPARQL\footnote{\url{https://www.w3.org/TR/sparql11-query/}} to obtain a set of \verb|(subj,rel)| pairs. 
    
\begin{figure*}
    \centering
    \includegraphics[width=\linewidth]{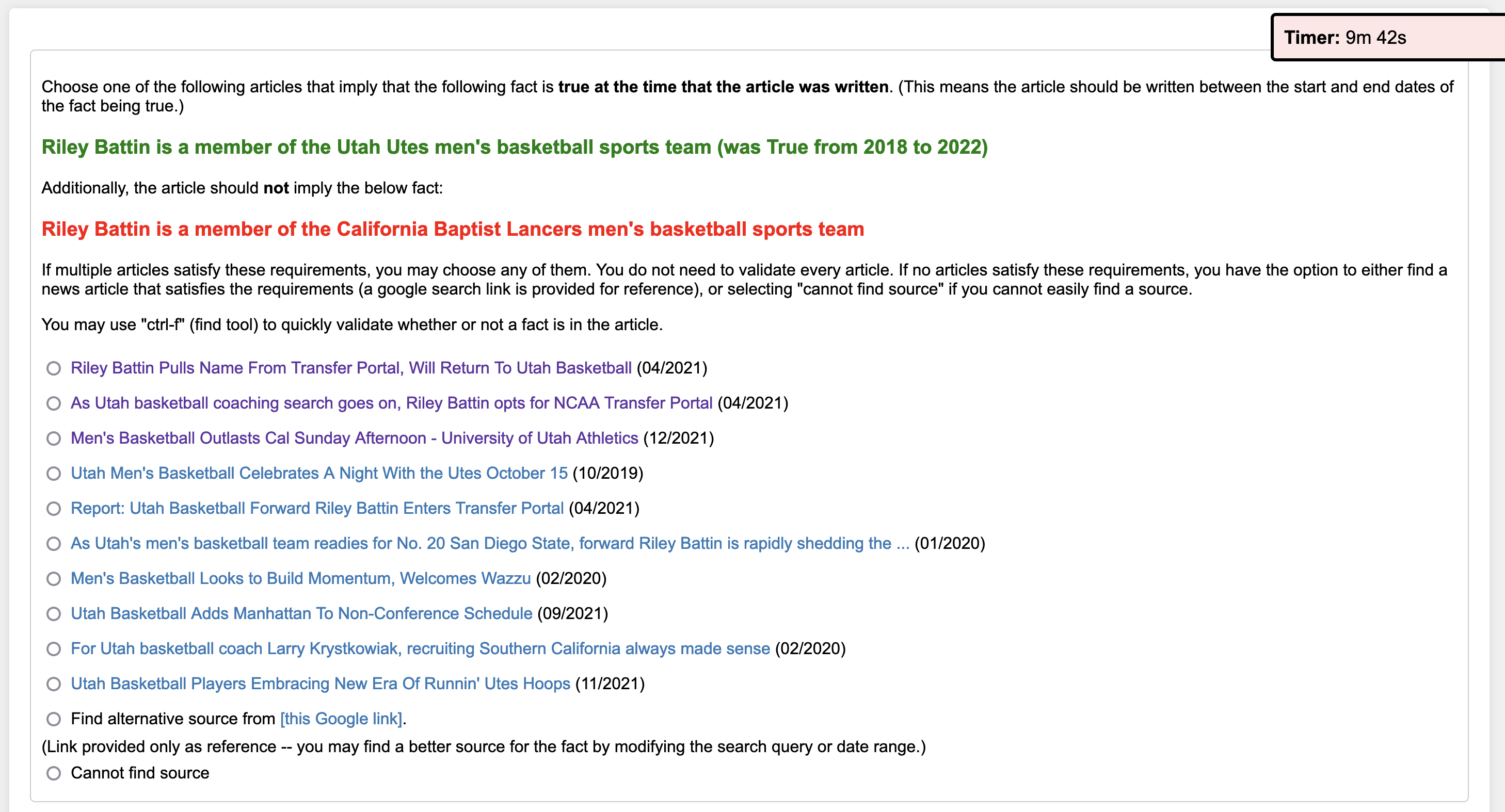}
    \caption{Screenshot of round 1 of annotation for news article.}
    \label{fig:annotator_screenshot_1}
\end{figure*}

\begin{figure*}
    \centering
    \includegraphics[width=\linewidth]{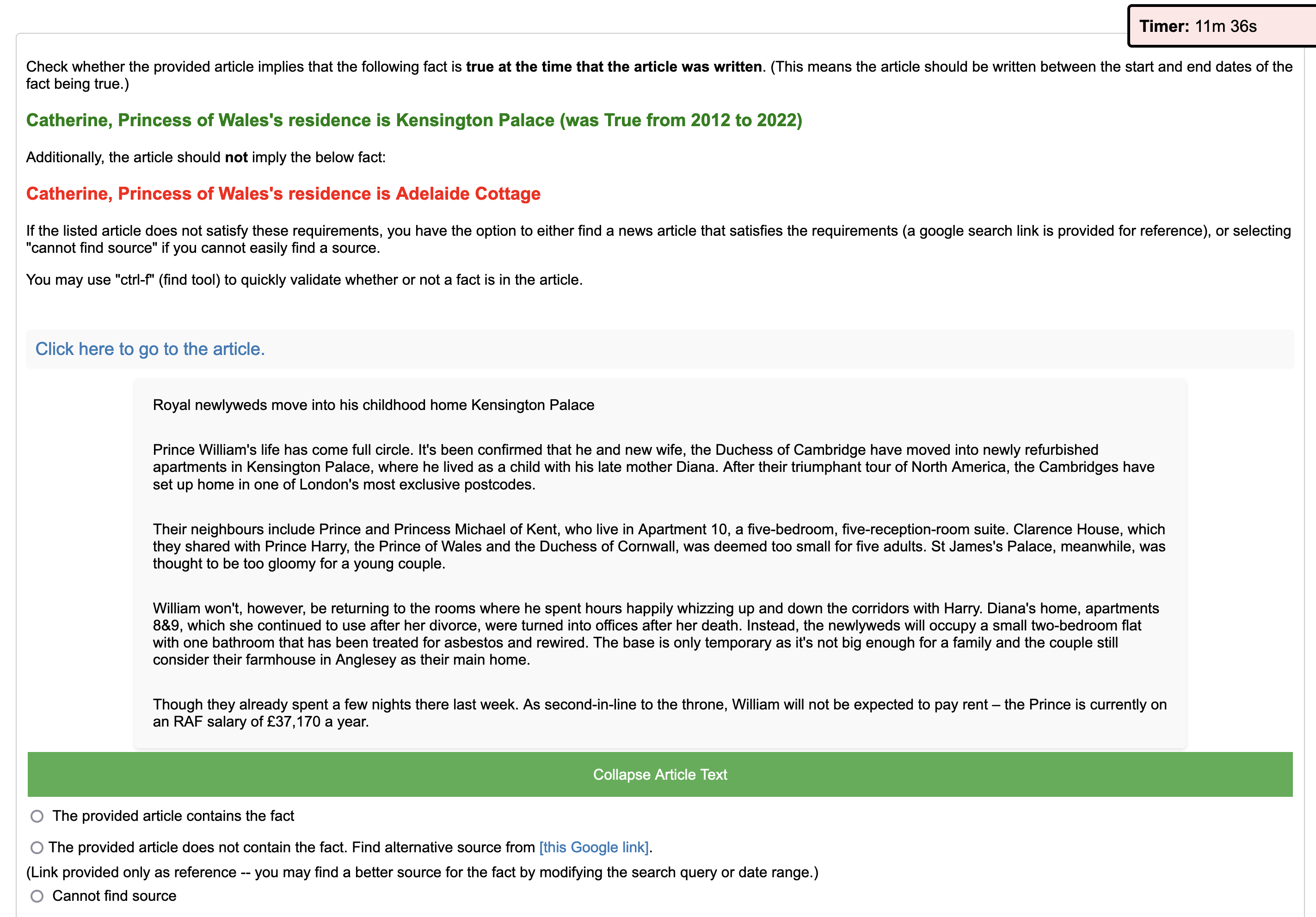}
    \caption{Screenshot of round 2 of annotation for news article.}
    \label{fig:annotator_screenshot_2}
\end{figure*}

\paragraph{Obtaining Documents $L$.} We annotate each timestamped relation, \verb|(subj,rel,obj,start_ts, end_ts)| with a source written between \verb|start_ts| and \verb|end_ts| (preferably close to the \verb|start_ts|) stating that the \verb|(subj,rel,obj)| relation is true. We crowdsource annotations from Prolific in two stages. In the first stage, Prolific annotators were presented with an interface which scraped candidate news articles off of Google\footnote{In particular, we set the to-be-matched parameter of the search to ``news'', i.e. \url{https://www.google.com/?tbm=nws}}, and were asked to select sources which stated that the fact \verb|(subj,rel,obj,start_ts, end_ts)| is true, but \textbf{did not} state that any succeeding fact, \verb|(subj,rel,obj2,start_ts2, end_ts2)| where \verb|start_ts2| $>$ \verb|start_ts|, is true.
In the second stage, we validated Prolific annotations from the first stage by presenting articles from the first round of annotations to annotators in the second round, and asking users whether those articles contained the fact in question. If second annotator does not affirm the fact is present in the article, we throw out the fact and the associated annotation.
\bzl{TODO: inter-annotator agreement}
We do an additional third round of filtration with a language model, asking the language model to affirm that the text of an article contains \verb|(subj,rel,obj,start_ts, end_ts)| but not any succeeding facts \verb|(subj,rel,obj2,start_ts2, end_ts2)|.
We only include articles and facts that pass all three rounds of annotation.
We recruited English-speaking participants from the US for annotations for all annotations. The full set of instructions we give annotators can be found in~\Cref{tab:annotator_instructions_1,tab:annotator_instructions_2}.
Screenshots of the interface can be found in~\Cref{fig:annotator_screenshot_1,fig:annotator_screenshot_2}.

\paragraph{Generating Question-Answers Pairs $(q,\{a\})$.} We automate generation of questions and answers from $W$ by writing templates for each relation and generating questions and answers from those templates. The full list of templates can be found in~\Cref{tab:wiki_qs_templates}.

\begin{table*}[]
    \centering
    \small
    \begin{tabular}{p{15cm}}
     \textbf{Please read these instructions carefully and only proceed once you have understood them. Once you start the task, you will have 10 minutes to get through as many questions as possible.}

    For each question, you will be presented a fact. Please find a news article that implies that the fact is true, according to the below requirements:
    \begin{enumerate}
    \item The article implies the fact, such that a reasonable person, without any prior knowledge, can infer that the fact is true from reading the article.

    Example: For fact Emad Mostaque is CEO of Stability AI (was True from 2020 to 2024-03-23)
        
    Good Sources: This startup is setting a DALL-E 2-like AI free, consequences be damned: Article says "...Stability AI CEO and founder Emad Mostaque wrote in a blog post"
    
    Bad Sources:  Artists can now opt out of the next version of Stable Diffusion: Cannot conclude fact from text of article

    \item The article is a news article or blog post.

    Example: For fact Taylor Aylmer is a member of the Racing Louisville FC sports tea

    Good Sources: Team News: Aylmer to make first regular season start

    Bad Sources: Taylor Aylmer - Racing Louisville FC Midfielder - ESPN, Taylor Aylmer - Instagram

    \item The fact is stated in the main body of the article text, not in a table, list, image, image caption, embedded tweet, etc.

    Example: For fact Taylor Aylmer is a member of the Racing Louisville FC sports team

    Good Sources: Team News: Aylmer to make first regular season start, Recap: Racing rallies to beat Orlando, keep playoff hopes alive: Fact is in a list at the end, not the main text

    Bad Sources: Jaelin Howell, Racing Louisville bring community together to help people with Down syndrome: Fact is in an image caption but nowhere in the main text

    \item The article is a web page, not a PDF or other file format.

    Example: For fact Ali Shojaie is a IMS Fellow

    Good Sources: Ali Shojaie elected fellow of the Institute of Mathematical Statistics 

    Bad Sources: IMS Carver Award 2023: Source is a PDF file, not a web page
    
    \item The article is written in English.

    Example: For fact Emad Mostaque is CEO of Stability AI (was True from 2020 to 2024-03-23)

    Good Sources: This startup is setting a DALL-E 2-like AI free, consequences be damned

    Bad Sources: [Bengali article]: Article is not in English
    
    \item Avoid articles that state that the fact is or is about to become false. These are generally written near or past the end date of a fact being true.

    Example: For fact Emad Mostaque is CEO of Stability AI (was True from 2020 to 2024-03-23)

    Good Sources: This startup is setting a DALL-E 2-like AI free, consequences be damned

    Bad Sources: Stability AI founder Emad Mostaque plans to resign as CEO, sources say: Article is about the fact being about to be false
    \end{enumerate}

If no listed articles satisfy these requirements, you have the option to either find a news article that satisfies the requirements (a google search link is provided for reference, you may need to manually adjust the query or date parameters) or selecting "cannot find source" if you cannot find any source in a reasonable amount of time.

There may also be a second fact that you need to avoid. If you see this fact in the article, do not select it as a source.

\textbf{Tip}: You may use "ctrl-f" (find tool) to quickly validate whether or not a fact is in the article.
    \end{tabular}
    \caption{Instructions for round 1 of annotation for news article.}
    \label{tab:annotator_instructions_1}
\end{table*}
\bzl{non-English not showing up!}

\begin{table*}[]
\small
    \centering
    \begin{tabular}{p{15cm}}
\textbf{Please read these instructions carefully and only proceed once you have understood them. Once you start the task, you will have 12 minutes to get through as many questions as possible.}

For each question, you will be presented a fact and a news article. Please confirm that the news article implies that the fact is true, and conforms to the below requirements:
\begin{enumerate}
    \item The article implies the fact, such that a reasonable person, without any prior knowledge, can infer that the fact is true from reading the article.

    Example: For fact Emad Mostaque is CEO of Stability AI (was True from 2020 to 2024-03-23)
        
    Good Sources: This startup is setting a DALL-E 2-like AI free, consequences be damned: Article says "...Stability AI CEO and founder Emad Mostaque wrote in a blog post"
    
    Bad Sources:  Artists can now opt out of the next version of Stable Diffusion: Cannot conclude fact from text of article

    \item The article is written in English.

    Example: For fact Emad Mostaque is CEO of Stability AI (was True from 2020 to 2024-03-23)

    Good Sources: This startup is setting a DALL-E 2-like AI free, consequences be damned

    Bad Sources: [Bengali article]: Article is not in English

    \item Avoid articles that state that the fact is or is about to become false. These are generally written near or past the end date of a fact being true.

    Example: For fact Emad Mostaque is CEO of Stability AI (was True from 2020 to 2024-03-23)

    Good Sources: This startup is setting a DALL-E 2-like AI free, consequences be damned

    Bad Sources: Stability AI founder Emad Mostaque plans to resign as CEO, sources say: Article is about the fact being about to be false
\end{enumerate}

If the provided article does not satisfy these requirements, you have the option to either find a news article that satisfies the requirements (a google search link is provided for reference, you may need to manually adjust the query or date parameters) or selecting "cannot find source" if you cannot find any source in a reasonable amount of time.

There may also be a second fact that you need to avoid. If you see this fact in the article, do not select it as a source.

\textbf{Tip}: You may use "ctrl-f" (find tool) to quickly validate whether or not a fact is in the article.

    \end{tabular}
    \caption{Instructions for round 2 of annotation for news article.}
    \label{tab:annotator_instructions_2}
\end{table*}

\begin{table*}[]
\small
    \centering
    \begin{tabular}{p{6cm}p{12cm}}
\multirow{2}{6cm}{
    \texttt{(\{subj\}, employer, \{obj\})}
} & \verb|Who is the employer of {subject}?| \\
& \verb|Is {subject} an employee of {object}?| \\
\multirow{3}{6cm}{
    \texttt{(\{subj\}, chief executive officer, \{obj\})}
} & \verb|Who is the CEO of {subject}?| \\
& \verb|What company is {object} the CEO of?| \\
& \verb|Is {object} the CEO of {subject}?| \\
\multirow{3}{6cm}{
    \texttt{(\{subj\}, chairperson, \{obj\})}
} & \verb|Who is the chairperson of {subject}?| \\
& \verb|What organization is {object} the chairperson of?| \\
& \verb|Is {object} the chairperson of {subject}?| \\
\multirow{3}{6cm}{
    \texttt{(\{subj\}, head of state, \{obj\})}
} & \verb|Who is the head of state of {subject}?| \\
& \verb|Where is {object} the head of state of?| \\
& \verb|Is {object} the head of state of {subject}?| \\
\multirow{2}{6cm}{
    \texttt{(\{subj\}, position held, \{obj\})}
} & \verb|What government position does {subject} hold?| \\
& \verb|Does {subject} hold government position {object}?| \\
\multirow{2}{6cm}{
    \texttt{(\{subj\}, member of sports team, \{obj\})}
} & \verb|What sports team is {subject} a member of?| \\
& \verb|Is {subject} a member of {object}?| \\
\multirow{3}{6cm}{
    \texttt{(\{subj\}, unmarried partner, \{obj\})}
} & \verb|Who is the unmarried partner of {subject}?| \\
& \verb|Who is the unmarried partner of {object}?| \\
& \verb|Is {object} the unmarried partner of {subject}?| \\
\multirow{2}{6cm}{
    \texttt{(\{subj\}, residence, \{obj\})}
} & \verb|Where does {subject} reside?| \\
& \verb|Does {subject} reside in {object}?| \\
\multirow{2}{6cm}{
    \texttt{(\{subj\}, headquarters location, \{obj\})}
} & \verb|Where is the headquarters location of {subject}?| \\
& \verb|Is the headquarters location of {subject} in {object}?| \\
\multirow{2}{6cm}{
    \texttt{(\{subj\}, P463, \{obj\})}
} & \verb|What organization is {subject} a member of?| \\
& \verb|Is {subject} a member of {object}?| \\
\multirow{2}{6cm}{
    \texttt{(\{subj\}, member of political party, \{obj\})}
} & \verb|What political party is {subject} a member of?| \\
& \verb|Is {subject} a member of {object}?| \\
    \end{tabular}
    \caption{Question-answer templates in the News domain}
    \label{tab:wiki_qs_templates}
\end{table*}

\paragraph{Prolific Details}

We recruited a total of 680 English-speaking prolific annotators from the United States, with each annotator spending an average of 16:50 minutes on the task ($\sim$ 7 minutes to read and understand instructions). We paid annotators an average of $\$14.20$ per hour.
This task was deemed exempt from IRB review.
No personally-identifiable information was collected or stored, and all prolific annotators were associated with an anonymous prolific ID.

\subsection{Synthetic Conversations}
\label{app:convos}
We also construct this dataset in three stages:
\paragraph{Generating World States $W$.} We model the underlying world and its transformations as a Markov chain with states $S$ and a set of allowable transitions $T(S)$ determined by $S$. At each timestep, we randomly sample a transition from $T(S)$ uniformly at random.
States $S$ are described by a set of relations \verb|(subj, rel, obj)|. 
The full list of entities types and relations for each entity type can be found in~\Cref{tab:convo_states}. To construct each world, we subsample 10 people and 5 companies, and randomly initialize their kinship and employment relations.
Transitions $t\in T(S)$ change one or more relation in the state. To be able to test the limits of our propagation, the set of transitions we define in this domain all change more than one relation: for example, ``\textit{Bob changed jobs to work at Google}'' changes the \textit{employees} of Google, the set of \textit{coworkers} of Bob, the set of \textit{coworkers} of all Google employees, and the set of \textit{coworkers} of all employees of Bob's former company, etc. The full list of transitions and their downstream effects can be found in~\Cref{tab:convo_transitions}.

\paragraph{Generating Conversations $L$.} We generate conversations by sampling two people in the world $p_1$ and $p_2$ and prompting two LLMs with their corresponding personas and initial facts.
We then generate twelve conversation ``chunks'' as follows:
We begin by sampling the next transition we want to make in the world.
The transition corresponds to a natural language string that corresponds to only a single relation. However, we know that each transition is associated with multiple changing relations. To be able to infer the \textit{downstream} changes of a single relation changing, we need to know auxiliary facts related to the \textit{object} of the changed relation.
In the multi-hop subset of this dataset, we mention auxiliary facts in the \textit{prior} conversation chunks, while only mentioning the immediate transition (on a single relation) in the current chunk (\textit{without} mentioning any downstream changes). Thus, to make the correct downstream inferences on this subset, the system must retrieve and reason across facts from prior conversation chunks.

For the singlehop subet, we mention \textit{all downstream effects} in the same conversation chunk that a transition is made.

\begin{table*}[]
    \centering
    \begin{tabular}{p{2cm}p{13cm}}
    \toprule
        Entity Type & Possible Relations \\
        \midrule
        Person & spouse, parents, children, job, company, hobbies, coworkers, work location, boss, salary, industry, is-employed-full-time, work hours, workplace, siblings, parents-in-law, children-in-law, step-parents, step-children, equipment necessary for hobbies \\
        Company & employees, jobs, head, location, industry, workplace type \\
        Job & company, salary, is-full-time, work hours \\
        Hobby & equipment necessary for hobby \\
        \bottomrule
    \end{tabular}
    \caption{Full list of entities and relations defining each world state in the Conversation domain.}
    \label{tab:convo_states}
\end{table*}

\begin{table*}[]
    \centering
    \begin{tabular}{p{4cm}p{10cm}}
    \toprule
        Transition type & Downstream effects \\ \midrule
        \texttt{person.job} changes from \texttt{job1} to \texttt{job2} & 
        person.company, person.coworkers, person.work-location, person.boss, person.salary, person.industry, person.is-employed-full-time, person.work-hours, person.workplace, job1.company.employees, job2.company.employees \\ \midrule
        \texttt{person.spouse} changes from \texttt{person1} to \texttt{person2} & person.parents-in-law, person.parents.children-in-law, person.children.step-parents, person.step-children, person1.spouse, person1.parents-in-law, person1.parents.children-in-law, person2.spouse, person2.parents-in-law, person2.parents.children-in-law, person2.children.step-parents, person2.step-children \\ \midrule
        \texttt{person} adopts \texttt{child} & person.children, child.parents, child.siblings, child.spouse.parents-in-law, person.children-in-law, child.step-parents, person.spouse.step-children, person.children.siblings \\ \midrule
        \texttt{person} gets a new hobby \texttt{hobby} & person.equipment-necessary-for-hobbies \\ \midrule
        \texttt{job.salary} changes & for all people that have that job: person.salary \\ \midrule
        \texttt{job.work-hours} changes & for all people that have that job: person.work-hours \\
        \bottomrule
    \end{tabular}
    \caption{Full list of possible state transitions in the Conversation domain. Note the set of available transitions may vary depending on the underlying state.}
    \label{tab:convo_transitions}
\end{table*}

\paragraph{Generating Question-Answers Pairs $(q,\{a\})$.} Given a world state at time $t$, we query \textit{all} facts about the world. Similar to the news setting, we automate generation of questions and answers through templates. Templates in this setting can be found in~\Cref{tab:convo_qs_templates}.

\begin{table*}[]
\small
    \centering
    \begin{tabular}{p{8cm}p{7cm}}
\multirow{2}{8cm}{
    \texttt{(\{subj\}, spouse, \{obj\})}
} & \verb|Who is the spouse of {subj}?| \\
& \verb|Who is the spouse of {obj}?| \\
\multirow{1}{8cm}{
    \texttt{(\{subj\}, job, \{obj\})}
} & \verb|What is the job of {subj}?| \\
\multirow{1}{8cm}{
    \texttt{(\{subj\}, company, \{obj\})}
} & \verb|Which company does {subj} work at?| \\
\multirow{1}{8cm}{
    \texttt{(\{subj\}, hobbies, \{obj\})}
} & \verb|List all known hobbies of {subj}.| \\
\multirow{1}{8cm}{
    \texttt{(\{subj\}, coworkers, \{obj\})}
} & \verb|List all known coworkers of {subj}.| \\
\multirow{1}{8cm}{
    \texttt{(\{subj\}, work location, \{obj\})}
} & \verb|In which city does {subj} work?| \\
\multirow{1}{8cm}{
    \texttt{(\{subj\}, boss, \{obj\})}
} & \verb|Who is the head of {subj}'s workplace?| \\
\multirow{1}{8cm}{
    \texttt{(\{subj\}, salary, \{obj\})}
} & \verb|What is the salary of {subj}?| \\
\multirow{1}{8cm}{
    \texttt{(\{subj\}, industry, \{obj\})}
} & \verb|What industry does {subj} work in?| \\
\multirow{1}{8cm}{
    \texttt{(\{subj\}, is-employed-full-time, \{obj\})}
} & \verb|Does {subj} work full-time or part-time?| \\
\multirow{1}{8cm}{
    \texttt{(\{subj\}, work-hours, \{obj\})}
} & \verb|What are the work hours of {subj}?| \\
\multirow{1}{8cm}{
    \texttt{(\{subj\}, workplace, \{obj\})}
} & \verb|What type of workplace does {subj} work out of?| \\
\multirow{1}{8cm}{
    \texttt{(\{subj\}, parents, \{obj\})}
} & \verb|List all parents of {subj}.| \\
\multirow{1}{8cm}{
    \texttt{(\{subj\}, children, \{obj\})}
} & \verb|List all children of {subj}.| \\
\multirow{1}{8cm}{
    \texttt{(\{subj\}, siblings, \{obj\})}
} & \verb|List all siblings of {subj}.| \\
\multirow{1}{8cm}{
    \texttt{(\{subj\}, parents-in-law, \{obj\})}
} & \verb|List all parents-in-law of {subj}.| \\
\multirow{1}{8cm}{
    \texttt{(\{subj\}, children-in-law, \{obj\})}
} & \verb|List all children-in-law of {subj}.| \\
\multirow{1}{8cm}{
    \texttt{(\{subj\}, step-parents, \{obj\})}
} & \verb|List all step-parents of {subj}.| \\
\multirow{1}{8cm}{
    \texttt{(\{subj\}, step-children, \{obj\})}
} & \verb|List all step-children of {subj}.| \\
\multirow{1}{8cm}{
    \texttt{(\{subj\}, necessary equipment for hobby, \{obj\})}
} & \verb|List all equipment {subj} needs for their hobbies.| \\
    \end{tabular}
    \caption{Question-answer templates in the Conversation domain}
    \label{tab:convo_qs_templates}
\end{table*}

\begin{table}[!t]
    \centering
    \footnotesize
    \begin{tabular}{lcc}
    \toprule
       Relation type  & \# \texttt{(s, r)} & \# \texttt{(s, r, o)} \\
       \midrule
       Member of sports team & 284 & 382 \\
       Position held & 164 & 382 \\
       Employer & 38 & 77 \\
       Chairperson & 20 & 42 \\
       Head of state & 9 & 18 \\
       CEO & 6 & 13 \\
       Unmarried partner & 5 & 12 \\
       Residence & 4 & 8 \\
       Headquarters & 2 & 4 \\
       Member of political party & 1 & 2 \\
       \midrule
       Total & 533 & 1174  \\
       \bottomrule
    \end{tabular}
    \caption{Breakdown of changed relation types in the News domain, categorized by number of unique \texttt{(subj, rel)} pairs and \texttt{(subj, rel, obj)} triples.}
    \label{tab:news_relations}
\end{table}

\section{Dataset Statistics}
\label{app:dataset_stats}
\begin{figure*}
        \centering
        \includegraphics[width=\linewidth]{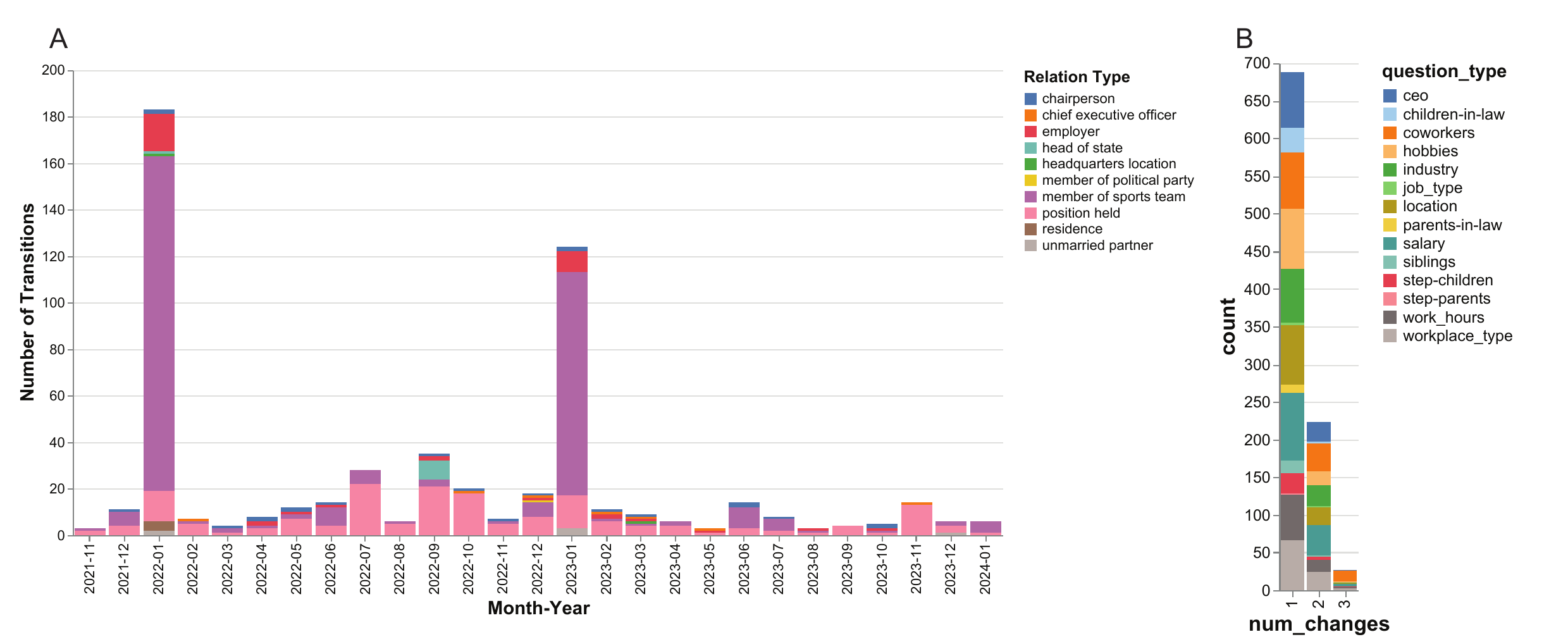}
        \caption{Distribution of changed relation types in the (A) News and (B) Conversation domains. (A) depicts the number of transitions per year of each month, while (B) depicts the number of relation types that underwent that number of changes across all conversations.}
        \label{fig:convos_stats}
\end{figure*}

The breakdown of changes in each of our datasets can be found in~\Cref{tab:news_relations} for news articles and~\Cref{fig:convos_stats} for conversations. The breakdown of questions for conversations can be found in~\Cref{tab:convos_questions}.

\begin{table}[!t]
    \centering
    \resizebox{\columnwidth}{!}{
    \footnotesize
    \begin{tabular}{lccc}
    \toprule
       Question Topic & \# Yes/No & \# Multiple Choice &  \# MC Choices \\
       \midrule
       Boss & 140 & 74 & 26\\
       Coworkers & 481 & - & -\\
       Industry & - & 74 & 26 \\
       Is employed full-time & 82 & - & -\\
       Salary & 158 & 80 & 11 \\
       Work hours & 110 & 64 & 10 \\
       Work Location & 274 & 72 & 20 \\
       Workplace & 140 & 74 & 26 \\
       \midrule
       Total & 1385 & 438  \\
       \bottomrule
    \end{tabular}
    }
    \caption{Distribution of generated questions in the Synthetic Conversation domain, categorized by question topic and type.}
    \label{tab:convos_questions}
\end{table}

\section{Qualitative Analysis}

\subsection{Error Analysis: Conversations (Multihop)}
\label{sec:multihop_errors}
Prototypical examples of multihop edit errors can be found below:

\paragraph{Retrieval Errors} Sometimes, the full list of facts that need to be updated are not retrieved. For example:

Input conversation chunk:
\begin{quote}
\it{
2023-11-01

Katie: Hey Olivia! How have you been? Guess what? I've changed my job to Library Assistant at Central Public Library! What's new with you?

Olivia: Hey Katie! That's amazing news! Congrats on the new job as a Library Assistant at Central Public Library!

...
}
\end{quote}

Retrieved facts:

\begin{itemize}
\it{
\item The role of General Practitioner at HealthFirst Medical Clinic is a full-time job.
\item The role of Library Assistant at Central Public Library is a full-time job.
\item Rachel has a full-time job as a Medical Assistant at HealthFirst Medical Clinic.
\item Rachel works at HealthFirst Medical Clinic as a Medical Assistant.
\item The work hours of a Library Assistant at Central Public Library are from 9 to 17.
\item Katie works full-time at the Urban Development Project.
\item Olivia works full-time at the airport.
\item The salary for a Library Assistant at Central Public Library is \$80,000.
\item Peter works from 9 to 17 at the Central Public Library.
\item Diana and Liam both have full-time jobs at the Urban Development Project.
\item The salary of an Archivist at Central Public Library is \$130,000.
}
\end{itemize}

The retriever at step 1 failed to retrieve the following facts to update. At inference time, these facts were retrieved and affected the LM's answer on questions:
\begin{itemize}
{\color{red}
\it{
\item Katie's working hours are from 10 to 15.
\item Katie's salary is \$100000.
\item Diana and Liam work with Katie.
}
}
\end{itemize}

\paragraph{Updating Errors}
Sometimes, facts that needed to be updated were retrieved, but were not updated by the LM. For example,

Input conversation chunk:
\begin{quote}
\it{
2023-11-26

Katie: Hey Olivia! How have you been? I heard you changed your job to Safety Officer at Urban Development Project! How's the new role treating you?

Olivia: Hey Katie! Yes, I did change my job to Safety Officer at Urban Development Project! The new role has been quite a change, but I'm really enjoying it so far. How have you been?

...
}
\end{quote}

Retrieved fact:

\begin{itemize}
\it{
\item Olivia works from 9 to 17 at her job.
}
\end{itemize}

Model-outputted update (and reasoning):
\begin{quote}
\begin{lstlisting}
Summary of changes:


    Olivia changed her job to Safety Officer at Urban Development Project.

    No information is provided about Olivia's work schedule.


Reasoning:

The input does not provide any information that would suggest Olivia's work schedule has changed. The conversation is about Olivia's new job, Katie's hobbies, and their plans for social activities. There is no mention of Olivia's work schedule or any changes that would affect it. Therefore, the fact "Olivia works from 9 to 17 at her job" is likely still true as of 2023-11-01.

Answer: No Change
\end{lstlisting}
\end{quote}

Note that this fact should've been flagged for potential updating and forwarded to the rewrite step (where the LM has an opportunity to retrieve more facts to confirm or deny this fact). 
However, the LM does not classify this fact as \textit{make false} because it is not directly contradicted by the input.

\section{Use of AI Assistants}
Code was written with Co-pilot turned on. GPT* models were also consulted for creating acronyms for the method and dataset names.

\end{document}